\documentclass[conference]{IEEEtran}
\usepackage{graphicx}
\usepackage{amsmath, amssymb, amsfonts}
\usepackage{cite, color}
\usepackage{algorithm}
\usepackage{algpseudocode}

\graphicspath{{./figures/}}

\newcommand{\reals}{\mathbb{R}}

\providecommand{\abs}[1]{\lvert#1\rvert}

\newcommand{\X}{\mathbf{X}}
\newcommand{\w}{\mathbf{w}}
\newcommand{\x}{\mathbf{x}}
\newcommand{\y}{\mathbf{y}}
\newcommand{\Y}{\mathbf{Y}}
\newcommand{\f}{\boldsymbol{f}}
\newcommand{\bt}{\boldsymbol{\beta}}
\newcommand{\cS}{\mathcal{S}}

\begin{document}

\title{Concept Drift Learning with Alternating Learners} %\\{\large\red (Edit the ELM section II.B to diversify the language + proof read + Paul's inputs  - 11am 11/15)}}


%\author{\IEEEauthorblockN{Yunwen Xu}
%\IEEEauthorblockA{Machine Learning Laboratory\\GE Global Research\\San Ramon, CA, USA\\yunwen.xu@ge.com}
%\and
%\IEEEauthorblockN{Rui Xu}
%\IEEEauthorblockA{Machine Learning Laboratory\\GE Global Research\\Niskayuna, NY, USA\\xur@ge.com}
%\and
%\IEEEauthorblockN{Weizhong Yan}
%\IEEEauthorblockA{Machine Learning Laboratory\\GE Global Research\\Niskayuna, NY, USA\\yan@ge.com}
%\and
%\IEEEauthorblockN{Paul Ardis}
%\IEEEauthorblockA{Machine Learning Laboratory\\GE Global Research\\Niskayuna, NY, USA\\ardis.p@ge.com}
%}

\author{\IEEEauthorblockN{Yunwen Xu\IEEEauthorrefmark{1},
Rui Xu\IEEEauthorrefmark{2},
Weizhong Yan\IEEEauthorrefmark{2} and
 Paul Ardis\IEEEauthorrefmark{2}}
\IEEEauthorblockA{\IEEEauthorrefmark{1} Machine Learning Lab\\
GE Global Research, San Ramon, CA, 94583\\ Email: yunwen.xu@ge.com}
\IEEEauthorblockA{\IEEEauthorrefmark{2}Machine Learning Lab\\
GE Global Research, Niskayuna, NY, 12309\\
Emails: xur@ge.com, yan@ge.com, ardis.p@ge.com}}

\maketitle

\begin{abstract}
Data-driven predictive analytics are in use today across a number of industrial applications, but further integration is hindered by the requirement of similarity among model training and test data distributions. This paper addresses the need of learning from possibly nonstationary data streams, or under {\em concept drift}, a commonly seen phenomenon in practical applications. A simple dual-learner ensemble strategy, {\em alternating learners} framework, is proposed. A long-memory model learns stable concepts from a long relevant time window, while a short-memory model learns transient concepts from a small recent window. The difference in prediction performance of these two models is monitored and induces an alternating policy to select, update and reset the two models. The method features an online updating mechanism to maintain the ensemble accuracy, and a concept-dependent trigger to focus on relevant data. Through empirical studies the method demonstrates effective tracking and prediction  when the steaming data carry  abrupt and/or gradual changes.
\end{abstract}


\section{Introduction}

%\begin{itemize}
%\item High level intro:popularity of statistical predictive models for industrial applications. The need: modeling under nonstationary environment, fast response to changes; robustness of models (window right after changes); incremental learning (not storing infinite data); suitable for efficient train and test 
%\item 
%\end{itemize}

\subsection{Background}

With the advancement in technologies of online data mining,  scalable and low cost data management, and distributed and cloud computing, data-based modeling for streaming data has gained tremendous popularity in industrial asset management. The use cases range from statistical process control to monitor machine operation condition for manufacturing systems \cite{montgomery2007introduction, macgregor1995statistical}, early failure detection and diagnosis as prognostics and health management \cite{macgregor1995statistical}, detecting and tracking objects and abnormalities in video streams \cite{parker2015detecting}, and monitoring and forecasting application demand to dynamically improve resource allocation utility \cite{ghanbari2011tracking}, just to name a few.  In many of these scenarios, the relationship between the modeling target and the available observations are often characterized using various machine learning models.  

Conventional machine learning models assume that the training data well represent the relationship in all  data drawn from the process of interest. %training data and test data follow the same distribution, 
In other words, data stationarity is required over the course of model learning and application. However, this is commonly violated in real practice, especially for applications involving streaming analysis. In many industrial use cases, data recorded from manufacture and operation processes demonstrate an inherited  nonstationary nature. For example, a sensor's measurement can drift due to the sensor's fault or aging,  changes in operation conditions or control command, and machine  degradation as a result of wearing. Even without seeing any drift in available data, there is no guarantee that data to become available  in the future follow similar distribution as the past in a stream setting. For all these observed, unseen, or uncertain changes in data, or in the process, we call them {\em concept drift} as in the literature.

On account of these practical limitations, the modeling technologies can either be used under constraints, or have to be adapted to handle nonstationarity. The first point of view follows the thought process of  dynamic systems modeling \cite{ljung1999system}. It assesses the robustness of a system model by carrying out sensitivity analysis and uncertainty decomposition, which results in a set of {\em feasible} working conditions (states) where models are valid and stable. Once the actual system states do not meet those constraints, the model does not provide any analytic conclusions. Seeing this disadvantage, most recent researches take the latter perspective, and continuously cope with the drifting concept. The problem of {\em concept drift learning} can be stated as: sequentially given (labeled) measured data that may be nonstationary over time, how to construct a predictive model that 1) predicts in a timely manner, and 2) tracks and responds to meaningful changes in data, and adapt to valid system behavior.

The term concept drift roughly denotes the source to generate {\em observations} $\x_t$ and {\em target} $\y_t$, or the joint distribution of the time-indexed random vectors $\cS_t = \{(\X_t, \Y_t)\}$ for $t = 1, 2, \dots$,   changes over time $t$ \cite{zliobaite2009learning}.  The causes can be categorized as: change in data distribution $P(\X_t)$, so that a fitted model that achieves optimal averaged prediction performance for one period of time (the training phase) is no longer optimal for the new test data; this is also called {\em virtual drift}, since the response $\y$ for a same $\x$ remains the same.  The other type of drift refers to the change in the conditional distribution $P(\Y|\X)$, describing the relationship between $\X$ and $\Y$. This  is also called {\em real drift}, since response may change despite the same observation. Changes purely caused by seasonal effect and operational condition (e.g., temperature, humidity) are examples of virtual drift; while machine degradation belongs to real drift. 

Depending on how model adapts to newer concepts, existing works can be generally classified as trigger-based and evolving methods. 
%Among the extensive researches to learn under nonstationary environment, trigger-based and evolving methods are two general approaches. 
The {\em trigger-based} approaches explicitly perform concept change detection on input streams, the derived feature space, or model performance metric; various window based, or sequential hypothesis tests have been developed,  aiming at an optimal detection rate and sample efficiency  \cite{alippi2008just1, wang2015concept, klinkenberg2000detecting}. Once a drift is detected, the models are retrained or incrementally updated with valid training data from recent experience or through resampling  \cite{harel2014concept, Bach08_pl, yang2006mining, Fan:2004:SDS:1014052.1014069}. In contrast, the {\em evolving} approaches continuously update the model without an explicit goal to detect change. Many online learning models are evolving over time. The most studied methods with overall better  performance utilize an ensemble of base models, different implementations are distinguished by the type of base models, training set composition methods, ensemble update rules, and the way model output is fused, \cite{soares2015dynamic, Katakis:2008:ECC:1567281.1567458, Kolter:2007:DWM:1314498.1390333, Street:2001:SEA:502512.502568}. It is commonly viewed that the trigger-based approaches work well for {\em abrupt changes}, while evolving approaches can pick up {\em gradual changes} with less delay, more discussions can be found in recent survey papers on this topic \cite{zliobaite2009learning}, \cite{ditzler2015learning},\cite{Gama2014_survey}. 

In addition to the challenges sharing across concept drift learning literature, the recent rapidly rising pull from the Industrial Internet of Things (IIOT) applications  also set further constraints on computational resources: some analysis are performed in embedded systems  with limited computing power and memory (edge analytics), and this sometimes makes typical ensemble learning technologies unsuitable.  

{\bf Notations:} Throughout this paper, we use regular lower case letters (e.g., $x$, $\delta$) to denote single scalar observation or parameter,  bold lower letters (e.g., $\x, \y$) to denote vectors or realization of random vectors, and upper case bold letters to denote matrices or random vectors  (e.g., $\X$, $\Y$). The subscripts means time stamp (e.g., $\x_t$, $\X_t$). 

% a practical challenge of streaming analysis is further complicated by the high sample rate streams in internet-of-things (IoT) application and the need of edge analytics: conduct computation in the embedded system, so both the training effort and storage space for typical ensemble learning technologies become a big burden. 

%In this work, we do not strictly distinguish the source and behavior of the drift, but focus on proposing a light-weight, easy to implement framework that can continuously adapt to new concept, and provide high quality predictions.  In particular, we propose a procedure that utilizes {\em alternating learners} that is capable to see the drift, update the knowledge towards the most relevant experience, then either retrain or incrementally adjust its predictive relationships. The idea of alternating learners is motivated by the work by Bach and Maloof \cite{Bach08_pl},  modified to add more robustness on noise environment and extended to suite better on regression cases. 

\subsection{Contribution}

In this paper, we propose a light-weight, easy-to-implement {\em alternating learner} framework that continuously adapt to new concept, and provide high quality predictions. In particular, we employ two models:  a {\em long-memory} learner (L) that is trained on a long time window (LW) with samples relevant to the current concept, and a {\em short-memory} (S) learner that is trained on a short window (SW) that only contains most recent samples. The intuition is: when the concept is stationary, L performs no worse than S, since the training data of S is a subset of L, and data distribution is homogeneous; however, on occurrence of concept drift, S adapts faster since the percentage of newer samples in SW is typically much higher than LW. This gives us the rule of alternating:
As a new batch of measurements and corresponding labels arrive, S and L are individually tested on the new samples. If performance of L is acceptable, then frontier of LW moves forward, and L is updated using information in the fresh samples. If L does not meet the performance target but S does, this is an indication of concept change. In this case, we truncate the tail of LW to be the same as SW, so all information prior to SW is retired. On the other hand, the fixed-sized SW window always slides forward with most updated information. As a result, we alternatively choose prediction from L and S.  And S  acts like a clutch to determine the boundary of each concept (LW), and serves as a (vague) change detector as well as providing a baseline after change. 

The idea is based on the work by Bach and Maloof \cite{Bach08_pl}. Comparing with  \cite{Bach08_pl}, we chose the extreme learning machine (ELM) as a base learner for regression cases, and
\begin{enumerate}
\item designed a new learner comparison criterion that better suits the regression case,
\item modified the learner alternating policy to prevent frequent reset under noisy environment, and
\item included data quality check to mitigate overfitting effect due to small sample size followed by learner reset.
\end{enumerate}
In  \cite{Bach08_pl}, extensive comparative studies have been done on the paired learner (PL) approach against most popular trigger-based and ensemble algorithms in classification setting, and  PL has outperformed or shown comparable performance with much less computational costs. So in this paper, we focus on comparison with the PL  algorithm and two other baseline models. Through a set of empirical studies, we show that the alternating learner approach successfully responses to both abrupt and gradual changes with different pattens, and provides the overall best prediction performance across comparison, and also achieves common industrial standard.

% {\red List of contributions goes here: }

The remainder  of the paper is organized as follows: %in Section \ref{sec:prelim}, we review the existing literature on concept drift treatment; 
in Section \ref{sec:method}, we describe  the alternating learner algorithm in detail, and propose an implementation using the online extreme learning machine as the base model; we then perform a case study on real power plant performance modeling and present the results  in Section \ref{sec:experiment}; 
%some case-based analysis and discussion on the proposed method is provided in  Section \ref{sec:discussion}; 
 we close the paper with conclusion  in Section \ref{sec:conclusion}.

\section{Method}
\label{sec:method}

\subsection{Alternating Learners Framework}
\label{al1}

% In this section, we describe the stragety of the alternating learner method. 

We consider a typical streaming analysis scenario: a set of labeled data (henceforth referred to as the {\em initial dataset}) are available in the beginning of learning process, after that, new measurements with labels become available one by one or in small batches (not necessary to have equal batch sizes). In industry practice, the initial dataset  is usually  from design of experiments (DoE), testbed runs or historical operation records. Sometimes they already contain a rich variety of concepts, but more often this is not sufficient due to limited duration and operation modes covered by the initial data. 

The  {\em alternating learners} (AL) framework consists of two learning models, L 	for long-memory learner and S for short-memory learner, an alternating policy is employed to choose the final prediction output either from L or S. 
% both are dynamically updated. 
Like many online sequential learning algorithms, the AL framework has two phases: {\em initial learning phase} and {\em continuous learning phase}.

In the initial learning phase, predictive models are built from the initial dataset. Without further assumption, this is a conventional (static) model training task. We initialize L trained by all initial dataset, and for S, at most $W$ samples from the latter portion (SW) are used, where $W$ is a user defined size  for the short memory window. Note that, only samples in SW will be kept in memory. We choose online learning algorithms to incrementally update L afterwards (except those reset instances), so there is no need to save an increasing volume of training data. The base model for L discussed in this paper is the online sequential extreme-learning machine (OSELM), which we will introduce in  subsection  \ref{elm}.

\begin{figure}
\begin{center}
\includegraphics[width=\columnwidth]{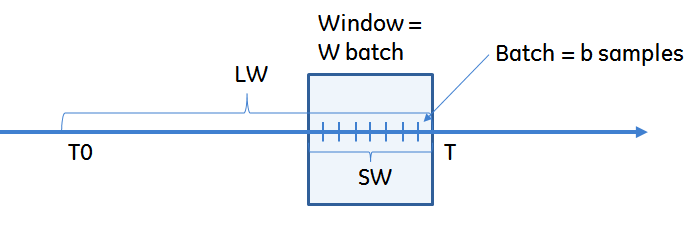}
\caption{The long-memory learner (L) captures the feature-response relationship over a window $[T_0, T]$ with relevant data, the short-memory learner (S) models ``current'' window of size $W$. $LW$ and $SW$ alternatively define the beginning of the learning window for stationary and changing environment respectively. } 
\label{fig-illustration}
\end{center}•
\end{figure}•

In the continuous learning phase, learning is driven by update, compare, and select the two models using the new available data, and concepts. The process is illustrated in Figure \ref{fig-illustration}. Model  S is always refit to the short window SW, serving as a baseline local prediction.  Model L continuously integrates new information by including new samples into its long-memory, $LW = [T_0, T]$ , where $T_0$ is the time since when new concept has not been observed. In case of concept change, outdated data must be discarded from training set by resetting $T_0$, and this is done by comparing predictive performance of L and  S, on the new data. 

Upon arrival of a new batch of $b$ samples $\{(\x_t, \y_t)\}_{t=T+1}^{T+b}$, where $\x_t \in \reals^d$ and $\y_t \in \reals^k$, the two learners both make prediction on the labels (response values). Let $\hat {\y}_t^L$ and $\hat {\y}_t^S$ be the prediction  and Err$(\hat {\y}_t^L)$ and Err$(\hat {\y}_t^S)$ be the corresponding performance of $L$ and $S$. The performance measure can be any error metric of interest, assuming the smaller the better, common choices can be the Mean Squared Error (MSE) or the Mean Absolute Percent Error (MAPE). The comparison outcome can be one of the following four cases:
\begin{enumerate}
\item Err$(\hat y_t^L)$  {\em is acceptable,} Err$(\hat y_t^S)>$  Err$(\hat y_t^L)$: there is no sign of concept drift, old data ($[T_0,T]$) is sufficient to train L to achieve the desired accuracy, performance of S is likely to be restricted by small training set;  
\item  Err$(\hat y_t^L)$ {\em is acceptable,} Err$(\hat y_t^S)\leq$  Err$(\hat y_t^L)$: although the local model shows a better performance on the new batch, since L still meets the prediction objective, this does not count towards a sign of concept drift. Moreover, in this case, we let L augment the training knowledge and pick up variability in system behavior; 
% ;  % there is no clear sign for concept drift, keep monitoring; 
\item  Err$(\hat y_t^L)$ {\em is unacceptable,} Err$(\hat y_t^S)>$  Err$(\hat y_t^L)$: predictions of both S and L are off. This may related to concept change, but a better substitution for L is not available; 
%if it appears after a stable period of Case 1 or Case 2. In this situation, this may occur right after an abrupt change, where both models fail to see the new data relationship.   
%If a new concept is ``real'', because  S adapts faster than L, this state will switch to Case 4; 
% if not purely random, can occur when the first  few data points corresponding to new concept enters the training window;  if the new concept is "real", since S adapts faster than L, this state will switch to Case 4; 
\item  Err$(\hat y_t^L)$ {\em  is unacceptable,} Err$(\hat y_t^S)\leq$  Err$(\hat y_t^L)$: local model outperforms long memory model in a meaningful way, this may indicate the distribution of pairs $(\x_t, \y_t)$ in the short window $[T-W, T]$ and the long window $[T_0, T]$ differs significantly, and model $S$ is closer to the upcoming concept.
% likely L is outdated, this occurs when the old data do not represent the current relationship. This possibility can be asserted if S further consistently outperforms L.
\end{enumerate}
From all four cases, only the last one both indicates a change and suggests a meaningful alternative for L. In other cases either performance of L is satisfactory (Case 1 and 2), or a better alternative is not available (Case 3). So we use the occurrence of Case 4 to design a  ``forgetting rule'' for  L. 

\begin{figure}
\begin{center}
\includegraphics[width=\columnwidth]{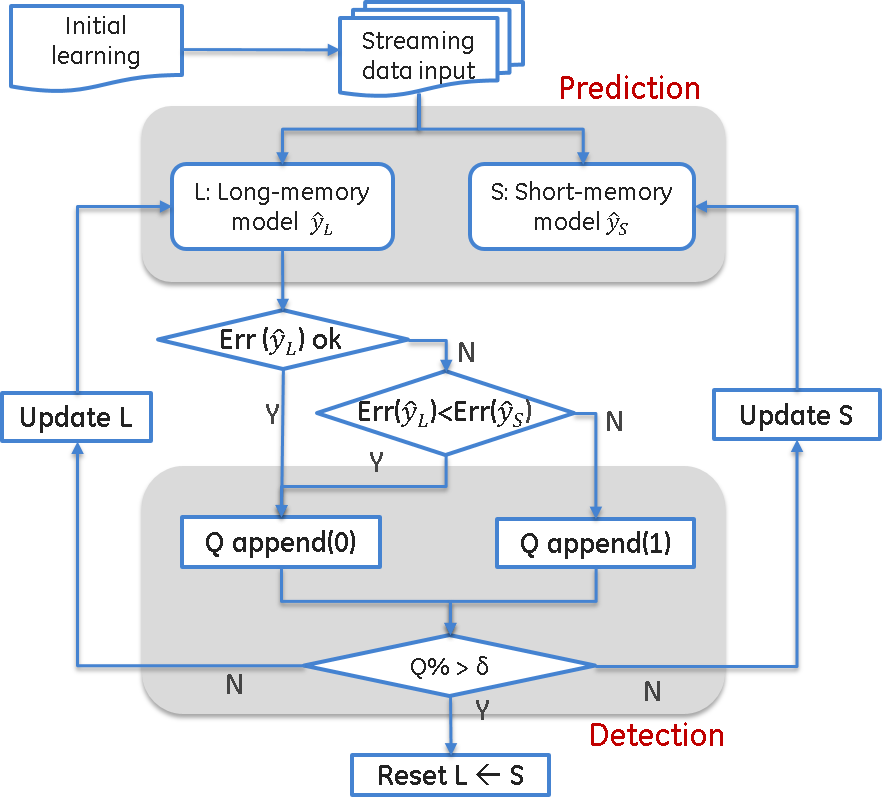}
\caption{The work flow of manage the alternating learners online. The prediction unit incrementally updates to learn from most recent data, the detection unit checks stationarity of data streams.} 
\label{fig-workflow}
\end{center}•
\end{figure}

In particular, we use a queue, Q, with a maximum length $W$ to record the status of performance comparison: register $1$ to Q whenever Case 4 happens, otherwise register $0$. The concept change is asserted when the rate of seeing $1$ is high in the Q, one criterion is the percentage of $1$ exceeds a user defined  threshold $\delta (0 < \delta <1)$, which related to forgetting sensitivity. When a new concept is asserted, we shrink window LW to match SW by setting $T_0 := T-W$, and also reset L by S, so that the performance of L after reset is not arbitrarily bad (warm restart). The flowchart of the algorithm is provide in Figure \ref{fig-workflow}. 

•

Overall, L represents a long and relative stable concept, and it aims at improving prediction accuracy by expanding training set. And S serves as a change detector, and also the alternative model to be exerted from an ensemble of two models. When concept is stable, L typically outperform S, and in the ensemble notion, we choose to output the prediction of L solely. When concept drift happen gently, L slowly catches up by online update. If change is rapid or the magnitude of the change is large, online update fails to follow since a large portion of training data are still associated to the old concept, then learners switch and reset mechanism determines  the beginning of new concept.

{\em Remarks:} 
\begin{itemize}
\item The user-specified reset threshold $\delta (0<\delta<1)$, the maximum length of the queue $W$, and the acceptable performance threshold $\tau>0$ are domain dependent  and can be pre-learned. In general, $\delta$ is not too small, small value results in frequent  reset, which leads to overall small training set size, and harms the overall prediction performance. We provide sensitivity study on $\delta$ and $W$ in Section \ref{sec:experiment}.

\item To prevent the alternating mechanism from overly reaction to noise, we also introduce a {\em leastWait} parameter $n_0$, so that reset is only allowed after the length of $Q$ exceeds $n_0$. In practice, $n_0$ is determined by reset threshold $\delta$ and sensitivity evaluation. For example, when $\delta \leq 0.5$, selecting $n_0 \leq 2$ essentially calls for a learner reset after seeing Case 4  one out of two checks after a previous reset. Frequent reset is undesired, especially when the learner accuracy fluctuation is caused by noise, since the information used to train the learners is almost as short as SW. In our  sensitivity experiments, we find for mid-range $\delta$ (between $0.3$ and $0.6$), performance does not change much with respect to $n_0$ as long as it is larger than $4$. 

\item We assume equal batch size $b$ for the ease of description, this framework also works for varying batch size, as long as it is compatible with the online updating algorithm. The OSELM algorithm we use for this study is able to handle varying batch size.  

\item The actual models to  implement  L and S are not restrictive: any model that can be updated in an online fashion (efficiently updated and does not store an increasing amount of data) will suffice.
\end{itemize}

\begin{algorithm}
\caption{Alternating Learners}\label{alg1}
\begin{algorithmic}[1]
\State {\bf Input:} $B_0$, $W$, $n_0$, $\delta$, $\tau$, $b$, $\{(\x_t, \y_t)\}, t = 1, 2, \dots, N$
\State $B_0$: number of batches available for initial training
\State $W$: maximum queue length to compare learners performance
\State $n_0$: minimum reset interval in batch count
\State $\delta$: reset threshold
\State $\tau$: threshold for acceptable performance 
\State $b$: number of samples per-batch

\Procedure{Alternating Learners}{$L$, $S$}%\Comment{$L$: long-memory learner, $S$: short-memory learner}
\State $T \gets B_0, T_0 \gets 0$
\State Train $L_1$, $L_2$, $S$ on initial dataset  $\{(\x_t, \y_t)\}_{t=1}^{Tb}$\Comment{$L_1$: complicated model requires many training samples,  $L_2$: simple model requires less training samples}
%\State $L_1$: complicated model requires many training samples,  %ELM, $k$ hidden neurons after cross validation
%\State $L_2$: simple model requires less training samples %linear regression,  $S$: ELM
\State $Q = []$ \Comment{Performance tracking queue}
	\While{$t<N:$} \Comment{New data come in}
	\State $\hat \y_t^{L1} \gets L_1.predict(\x_t)$, $\hat \y_t^{L2}\gets L_2.predict(\x_t)$
	\State $\hat \y_t^{S} \gets S.predict(\x_t)$
	%\If {$(T-T_0)b >= k$} \Comment{Choose long-memory learner}
	\If {not overfit} \Comment{Choose long-memory learner}
		\State $\hat \y_t^L \gets \hat \y_t^{L1} $
	\Else
		\State $\hat \y_t^L \gets \hat \y_t^{L2} $
	\EndIf

	\If {Err$(\hat \y_t^L) < \tau$ {\bf or} Err$(\hat \y_t^L) < $ Err$(\hat \y_t^S)$} 
		\State Q.append($0$)
	%\ElsIf {Err$(\hat y_t^L) < $ Err$(\hat y_t^S)$}
		%\State M.append($0$)
	\Else
		\State Q.append($1$)
	\EndIf

	\If {$len(Q)>W$} 
		\State Q.pop()
	\EndIf
	\If {$len(Q)>n_0$ {\bf and} $sum(Q)/len(Q) > \delta$ }  \Comment{Alternating condition}
		\State $T_0\gets T-W$
		\State $L\gets S$
		\State $Q \gets []$
	\EndIf
	
	\State Update $L_1$, $L_2$, $S$ using  $\{(\x_t, \y_t)\}$
	\EndWhile
\EndProcedure

\end{algorithmic}
\end{algorithm}

%
%{\em Remarks:} 
%\begin{itemize}
%\item We assume equal batch size $b$ for the easy of description, this framwork also works for varying batch size, as long as the prediction performance calculation is redefined.  
%\item The parameter $n_0$, or the minimum reset interval, defines the minimum number of batches between two resets, this is defined to prevent random reset before confirm on drift.
%\item The user-specified reset threshold $\delta, (0<\delta<1)$, the maximum length of the queue $W$, and the acceptable performance threshold $\tau>0$ are selected by intuition and domain understanding. In general $\delta$ is not too small, small value results in frequent  reset, which leads to overall small training set size, and therefore harm the overall prediction performance. We provide sensitivity study on $\delta$ and $W$ in Section \ref{sec:experiment}.
%
%\item The actual models to  implement  L and S are not restrictive: any model that can be updated in an online fashion will suffice. 
%\item In addition to flowchart, in implementation, we essentially maintains two versions of L: $L_1$ learns complicated concepts and $L_2$ learns simple hypothesis, we also add an {\em overfit test} for $L_1$, based on which we switch L to point to one of them. In our experiment, we  train online version of extreme learning machine as $L_1$, and overfit judgment is to check whether the number of training sample is larger than two times the hidden neuron number. We use linear regression model for $L_2$.  
%\end{itemize}•

\subsection{Base model selection: OS-ELM}
\label{elm}

We adopt the Extreme Learning Machine (ELM) as the base learners for L and S. The ELM is a special type of feed-forward neural networks introduced by Huang, et al. \cite{huang2006extreme}. Unlike in traditional feed-forward neural networks where training the network involves finding all connection weights and bias, in ELM, connections between input and hidden neurons are {\em randomly generated} and {\em fixed}. %This implies the output values from the hidden nodes are {\em known} for any given inputs,
This implies for any given inputs, the hidden neuron outputs are uniquely {\em determined};  this mapping and the values are referred to as {\em ELM feature mapping} and {\em ELM features}.
%that is, they do not need to be trained. 
Thus, training an ELM reduces to finding connection weights from the ELM features to output neurons only. %, which is simply a linear least squares problem whose solution can be directly generated by the generalized inverse of the hidden layer output matrix \cite{huang2006extreme}.  
Because of such special design of the network, ELM training becomes very fast. Numerous empirical studies and recently some analytical studies have shown that ELM has better generalization performance than other machine learning algorithms including SVMs and is efficient and effective for both classification and regression \cite{huang2006extreme, huang2014insight}.

Consider a set of $M$ training samples, $\{(\x_i,\y_i )\}_{i=1}^M, \x_i \in \reals^d, \y_i \in \reals^k$. Assume the number of hidden neurons is $K$. The output function of ELM for generalized SLFNs is
\begin{equation}\label{batch-elm}
\f (\x) = \sum_{i=1}^K \mathbb{\bt}_i h_i(\x) = {\bf H} (\x) \bt,
\end{equation}
where $h_i (\x)=G(\w_i, \omega_i, \x)$ is the output of $i^{th}$ hidden neuron with respect to the input $\x$,  and $\w_i \in \reals^d, \omega_i \in \reals$ are the neuron's weights and bias respectively; $G(\w,\omega,\cdot): \reals^d \mapsto \reals$ is a nonlinear piecewise continuous function satisfying ELM universal approximation capability theorems \cite{huang2014insight}, for example, the sigmoid function; $\bt_i$  is the output weight matrix between $i^{th}$ hidden neuron and the $k\geq 1$ output nodes. The ${\bf H}(\cdot)=[h_1 (\cdot), \dots, h_K (\cdot)]: \reals^d \mapsto \reals^K$  is the random feature map transforming a $d$-dimensional input to a $K$-dimensional random ELM feature.

For batch ELM where all training samples are available at the same time, the output weight vector $\bt\in \reals^{K \times k}$ can be estimated as the least-squares solution of $\boldsymbol{H \beta}=\boldsymbol{Y}$, that is, 
\begin{equation} 
\label{eq:elm}
\hat \bt = {\bf H}^\dagger {\bf Y}
\end{equation} where ${\bf H}^\dagger$ is the Moore-Penrose generalized inverse of the hidden layer output matrix (see \cite{huang2006extreme} for details), which can be calculated through orthogonal projection method: ${\bf H}^\dagger = (\bf H ^T \bf H)^{-1} {\bf H}^T$ when $\bf H$ has full column rank. 

The online sequential ELM (OSELM), proposed by Liang, et al. \cite{liang2006fast}, is a variant of classical ELM, it addresses the need of incrementally learning the output weights from chunks of data available in sequence. 
% which has the capability of learning data one-by-one or chunk-by-chunk with a fixed or varying chunk size. 
As described in details in\cite{liang2006fast}, OS-ELM involves two learning phases, {\em initial training} and {\em sequential learning}.

{\bf Initial training:} choose a small chunk of initial training samples,  $\{(\x_i,\y_i )\}_{i=1}^{M_0}$, where $M_0\geq K$, from the given $M$ training samples; and  calculate the initial output weight matrix, $\hat \bt^0$, using the batch ELM formula \eqref{eq:elm}. And initialize ${\bf R}_0 = ({\bf H}_0^T {\bf H}_0)^{-1}.$

{\bf Sequential learning:}  for the $(M_0+m+1)^{th}$ training sample, where $m = 1, \dots, M-M_0-1$, iterate between the following two steps. 
\begin{enumerate}
\item Calculate the partial hidden layer output matrix  for the new sample:
$$
{\bf H}_{m+1} = [h_1(\x_{M_0+m+1}), \dots, h_L(\x_{M_0+m+1})],
$$
and set the target ${\bf t}_{m+1} = \y^T_{M_0+m+1}$.
\item Calculate the output weight matrix:
\begin{equation} \label{oselm1}
\hat\bt^{m+1} =\hat \bt^m + {\bf R}_{m+1} {\bf H}_{m+1} ({\bf t}_{m+1}-{\bf H}^T_{m+1}\hat\bt^m)
\end{equation}
where
\begin{align} \label{oselm2}
%{\bf R}_{k+1}= {\bf R}_{k}-\frac{{\bf R}_{k} {\bf H}_{k+1} {\bf H}_{k+1}^T {\bf R}_{k}}{1+ {\bf H}_{k+1}^T {\bf R}_{k} {\bf H}_{k+1}}
&{\bf R}_{m+1}= \\ \nonumber 
&{\bf R}_{m}- {\bf R}_{m} {\bf H}_{m+1} ({\bf I}+ {\bf H}_{m+1}^T {\bf R}_{m} {\bf H}_{m+1})^{-1} {\bf H}_{m+1}^T {\bf R}_{m}.
\end{align}•
%for $m = 0, 1, 2, \dots, M-M_0+1$.
\end{enumerate}•
Note that the number of hidden node $K$ and the  random feature mapping is fixed after the initial training, only the ELM feature matrix ${\bf H}$ is re-calculated for new data chunks, and the output weight matrix $\hat\bt$ is updated accordingly. 

\subsection{Implementation}

The OSELM algorithm enjoys the benefits of fast training and incrementally updating, which fulfills our needs to maintain the long-memory learner L. For given (fixed) input-hidden layer weights $\{\w_i, \omega_i\}, i = 1, \dots, K$, the iteratively calculated output connection weight matrix $\bt^m$ by \eqref{oselm1} and \eqref{oselm2} is the same as the batch-mode solution when all $M_0+m$ training samples are fed at once.     The structure of the SLFN, or the number of hidden neurons K, is determined by performing a $k$-fold cross validation in initial learning phase. This structure parameter, $K$, once determined, remains the same for the online updating process. 

The short-memory learner, S, on the other hand, is repeatedly trained by batch mode ELM using all samples in SW, and can be served as the starting point of OSELM after switching. Note that, except the samples in SW, old data are not saved, so L is not completely retrained.

When solving for the output weights $\bt$ from SLFN, if  the number of sample points is less than number of hidden neurons, Equation \eqref{eq:elm}  become under-determined, and the fitted network is easily overfit. So in implementation, we maintain two versions of L: $L_1$ learns complicated concepts trained and updated by OSELM;  $L_2$ learns simple hypothesis, and we choose a linear regression model based on the same input features.  We add an {\em overfit test} for $L_1$, checking if the training samples in LW is more than $2K$,   we then switch the output of L to point to either $L_1$ or $L_2$. The detailed implementation steps are provided in Algorithm \ref{alg1}.

Here we make a note that, the comparison aims to conclude relative performance of the concept-drift handling frameworks, either triggering policies or ensemble rules, rather than the selection of base learner. Our base learner selection only considers the suitability of online update and efficient training/application.

\section{Experimental Results}
\label{sec:experiment}

We perform empirical performance assessment for the alternating learner algorithm in the regression setting.  % The experiments use time series data representing operation of a power gas turbine, which often demonstrates  concept drift phenomena. To cover various change behavior, we 
To demonstrate on an industrial use case that is  typically suffered by the issue of concept drift, we use simulated data  representing a power gas turbine operation with synthesized deterioration and maintenance work. The synthesized drift information we used in data generation are treated as ground truth. Comparison against three baselines are provided. 

{\bf Algorithms to compare with:} 
The conventional static machine learning model is implemented with the basic regularized ELM trained on the initial batch $B_0$ and tested on the remaining time series, this model misses the target right after concept change as expected. The OSELM is an example of typical online learning methods, under nonstationary environment, online learning gradually picks up new relationship but tends to have long lag, since the sensitivity and stability trade-off is  controlled globally. The third algorithm we implemented is the Paired Learner algorithm proposed in \cite{Bach08_pl}, the original algorithm is designed for classification, we re-implemented two versions for regression with slightly different queue registration criteria. Only results with better performance is  reported in this paper. 
%Online learning, full memory trains on all prior examples (no detection), Learning on fixed rolling window ($W$, window size), biased reservoir sampling (decay sample weights), SEA, WSEA. We set long-memory learner's reset trigger as $\delta = 0.5$, the minimum wait time $n_0 = 5$, and acceptable threshold for MAPE $\tau = 5$. Also Paired learners proposed, no minimum response window and 

{\bf Performance measure:} In multi-variate time series regression for power plant applications,  we choose the Mean Absolute Percent Error (MAPE) of the model prediction as the evaluation metric, i.e., for the $t^{th}$ batch, 
$MAPE(\hat\y_t, \y_t) = \frac{1}{b}\sum_{i=1}^b(\abs{\frac{\hat\y_i- \y_i}{\y_i}}\times 100\%)$, where $b$ is the batch  size. Note that, we assume the true response values ($\y_t$) of every new batch is available after $L$ and $S$ predict.

\subsection{Case study: gas turbine performance modeling for power plant}

In this case study, we consider to model the performance of a gas turbine in a power plant. The data are time series sensor readings from a power plant. A collection of  $9$ signals related to turbine operating status are simultaneously recorded:  compressor inlet temperature, compressor inlet humidity, ambient pressure, inlet pressure drop, exhaust pressure drop, inlet guide vane angle, fuel temperature, compressor flow, and controller calculated firing temperature. % In real plant,  generated power and other real time performance metrics are also measured, but for business sensitivity and the need to cover various change type, we use a gas turbine simulator to generate output. 

The generated power depends on the machine condition, and is comprehensively captured by a hidden variable {\em compressor efficiency} $\eta$.  The recorded sensor data and predefined compressor efficiency profile are fed into the GE internal power simulation tool, Gas Turbine Performance (GTP) to get the instantaneous gross electric power and net heat rate. The GTP simulator has been independently developed by GE Power, it has high fidelity, and requires large computational resources to configure and run. 

In real plant, the efficiency $\eta$ drops as regular wear, dust cumulation, coating material oxidation and some non-ideal usage condition. On an opposite side, maintenance activities such as cleaning and renovation improves machine status, and $\eta$ significantly increases after these events. In order to investigate the algorithm performance under different  types of concept drift. We generate different change patterns of $\eta$ and simulate turbine response respectively. Our learning target is to predict generated power using measured signal only.

To obtain statistical description of the algorithm performance, we generate $500$ datasets, each set is a sequence of $2000$ samples with inputs and outputs introduced above. We generate two change patterns of $\eta$:
\begin{itemize}
\item{\bf Abrupt change:} this pattern contains two sudden jumps on the compressor efficiency $\eta$. $\eta$ is set to be $1.0$ in the beginning, and linearly drops to $0.9$ in $l_1$ sample range, then suddenly jumps to $\eta = 1.1$. Then in $l_2$ sample range, $\eta$ slowly drops to $0.9$, and again jumps back to $1.1$. After that, $\eta$ stays around $1.1$ for $l_3$ sample range, then  begins to decrease, and finally ends at $0.95$ in the end of simulation. The duration of each section,  $l_1$, $l_2$ and $L_3$ are {\em randomly generated} positive integers satisfying $l_1+l_2+l_3 < 2000$.   
\item {\bf Gradual change:} this pattern contains one relatively long linear change of $\eta$. In this case, $\eta$ begins at  value $\eta=1.1$ and stays near constant for a duration of $l_1$ samples. Then $\eta$ linearly decreases to $0.9$ in $l_2$ sample range, after that it stays near $0.9$. Similarly, the random integers satisfies $l_1+l_2 < 2000$.  
\end{itemize}•
For both cases, independent and identically distributed Gaussian noise is added to every simulated efficiency profile and lasts for  the entire $2000$ time steps.   The dataset contains $265$  cases with abrupt change and $235$ with gradual change. 

\subsection{Performance summary}

We select the initial dataset with $B_0=100$ samples, and perform a repeated  $5$-fold  cross validation on each time series to select the best hidden node number for its  SLFN. 

\begin{figure}
\begin{center}
\includegraphics[width=\columnwidth]{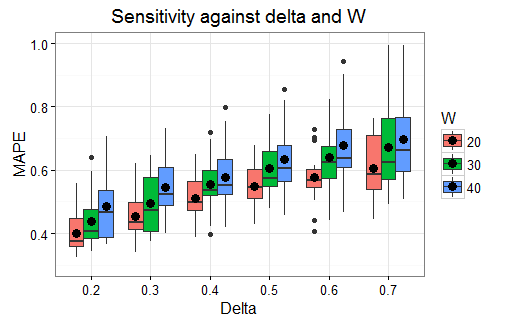}
\caption{Sensitivity of the prediction accuracy with respect to the alternating threshold, $\delta \in [0.2, 0.7]$ and short-memory window, $W \in \{20, 30, 40\}.$} 
\label{fig-sensitivity}
\end{center}•
\end{figure}•

The three user defined parameters are $W$, maximum queue length to compare learner performance; $\delta$, the learner switching threshold, and $n_0$, the least queue length allows a reset. As commented in Section \ref{al1}, $n_0$ and $\delta$ are closely related, once the threshold $\delta$ is given, we can choose $n_0$ properly, so we first determine the $\delta$ and  $W$ jointly by a preliminary sensitivity experiments: on a subset of $20$ randomly selected dataset from all the $500$ ($10$ abrupt cases and $10$ gradual cases). The prediction MAPE distribution on the $20$ sampled datasets are displayed in Figure \ref{fig-sensitivity}, the Box-and-Whisker plots indicate the average, median, and the $25$th and $75$th quantiles and extreme value ranges (end of vertical lines). It is seen that all combination show good MAPE score from $0.4\%$ to $0.8\%$, with an overall increasing trend with respect to $\delta$ and $W$. We choose $W = 20$, and $\delta = 0.4$. The reason not selecting $\delta = 0.2$ or $0.3$ is that, in order to avoid ``single instance reset'' described in Section \ref{al1},  selecting smaller $\delta$ leads to $n_0 > 7$, which is about $30\%$ of the queue length, and we decide to trade this toward response speed. 

\begin{figure}
\begin{center}
\includegraphics[width=\columnwidth]{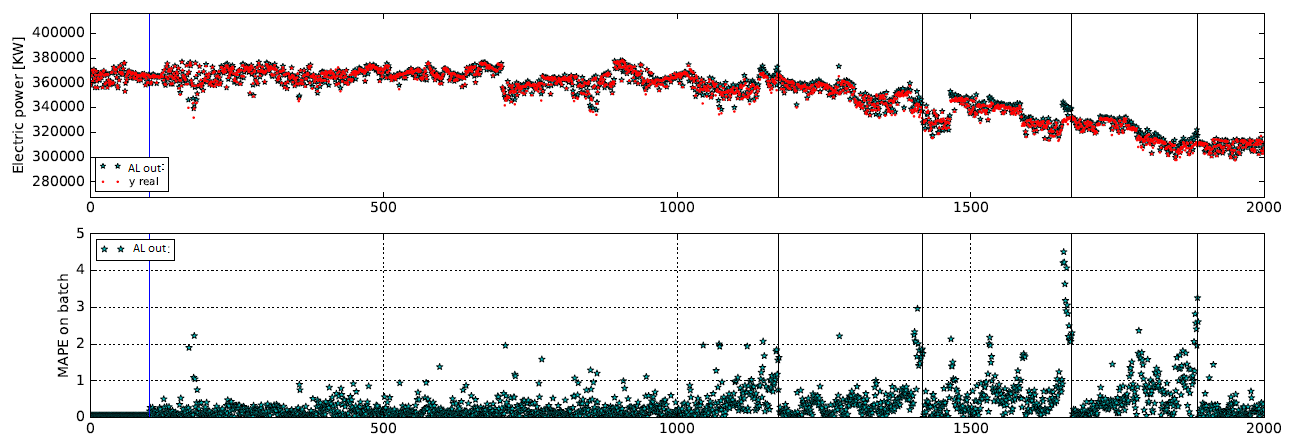}
\caption{An example with {\em gradual concept drift}: (Top) real (red) and predicted (green) electric power generated by the turbine, length of time series is $2000$. (Bottom) The batch-wise Mean Absolute Percent Error (MAPE) from the alternating learner output. The black vertical bars indicate the switching moment. Parameters: $B_0 = 100, b = 1, W = 20, \delta = 0.4, n_0 = 5$, average MAPE over time is 0.364\% . }
\label{ts-gradual}
\end{center}•
\end{figure}•

\begin{figure*}
\begin{center}
\includegraphics[width=1.8 \columnwidth]{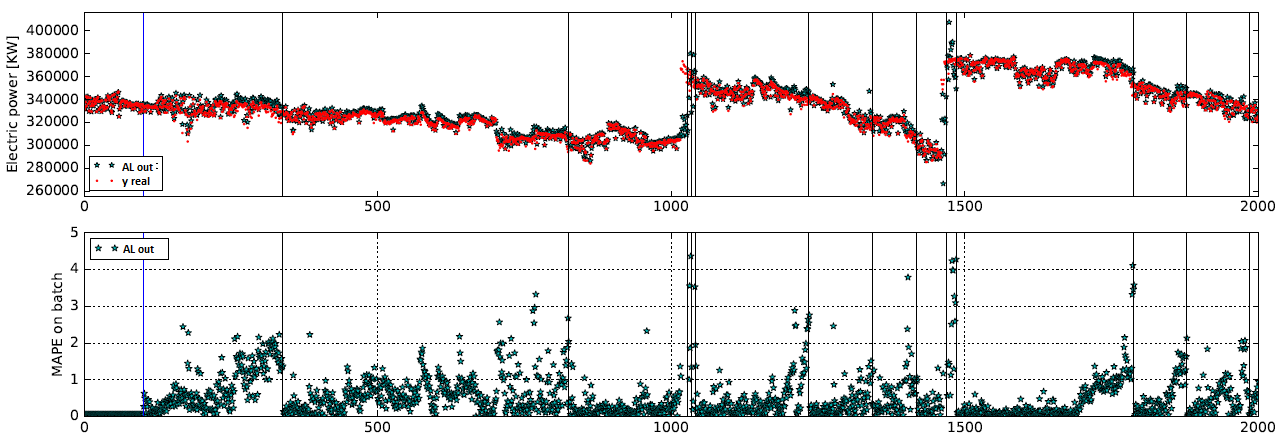}
\caption{An example with {\em abrupt concept drift:} (Top) real (red) and predicted (green)  electric power generated by the turbine, length of time series is $2000$. (Bottom) The batch-wise Mean Absolute Percent Error (MAPE) from the alternating learner output. Parameters: $B_0 = 100, b = 1, W = 20, \delta = 0.4, n_0 = 5$, average MAPE over time is 0.682\%. The blue and black vertical lines indicate beginning of online learning phase and the learner alternate instances. }
\label{ts-abrupt}
\end{center}•
\end{figure*}•

Examples of the algorithm prediction results are shown in Figure \ref{ts-gradual} (gradual) and Figure \ref{ts-abrupt} (abrupt). Assume data comes one sample at a time, so batch size $b=1$, $100$ batches are  available during the initial learning.  We take industrial standard and set the  performance acceptance limit $\tau = 1\%$. The figures show results when  selecting the maximum queue capacity as $W=20$ batches, and reset trigger to be $\delta=0.4$. On the top figures, real power output value is plotted in red, and the green stars are the alternating learner's prediction. The bottom plot depicts MAPE of prediction on each prediction batch (sample). Blue vertical line indicates the beginning of continuous learning phase, and the black lines indicate the time instance to reset long-memory learner.  It is seen that, reset usually occurs when the MAPE score shows an increasing trend. It happens less frequently  for gradual changes, and  can be caused by either slow concept drift, or local fluctuations (Figure \ref{ts-gradual}). The algorithm also responses quickly after abrupt changes, and the prediction performance returns to small MAPE range  ($1\%$) after the reset (Figure \ref{ts-abrupt}).

%\begin{figure}
%\begin{center}
%\includegraphics[width=\columnwidth]{boxplot_500.png}
%\caption{The distribution summary of the mean absolute percent error of the alternating learners on $500$ sets of simulated data steams. Each boxplot corresponds to a parameter set $W$: short window size, {\em theta}: L reset threshold.  The crosses at the center of boxes indicate the mean and lines indicate the median. } 
%\label{fig-procedure}
%\end{center}•
%\end{figure}•

%
%\begin{figure}
%\begin{center}
%\includegraphics[width=\columnwidth]{alg_comparison.png}
%\caption{The Box-and-Whisker plots of the prediction MAPE when the four algorithms perform on a set of $500$ simulated data streams. The red boxes represent the MAPE distributions when the data streams carry abrupt concept changes, and the green boxes represent the same when gradual changes are assumed. Black dots inside the boxes indicate the mean values of the distribution and the center bars of the boxes indicate the median values. } 
%\label{fig-comparison}
%\end{center}•
%\end{figure}•

\begin{figure}
\begin{center}
\includegraphics[width=.8\columnwidth]{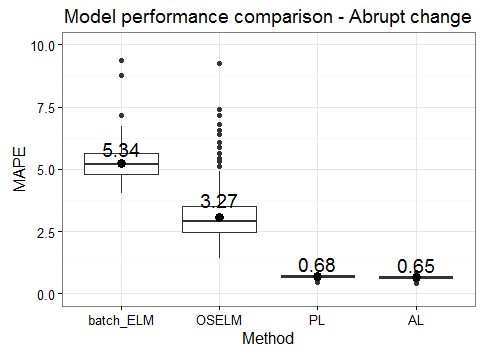}
\caption{The Box-and-Whisker plots of the prediction MAPE when the four algorithms perform on a set of $265$ simulated data streams with {\em abrupt}  concept drift. The black dots inside the boxes indicate the mean values of the distribution and the center bars of the boxes indicate the median values. } 
\label{fig-comparison-a}
\end{center}•
\end{figure}

\begin{figure}
\begin{center}
\includegraphics[width=.8\columnwidth]{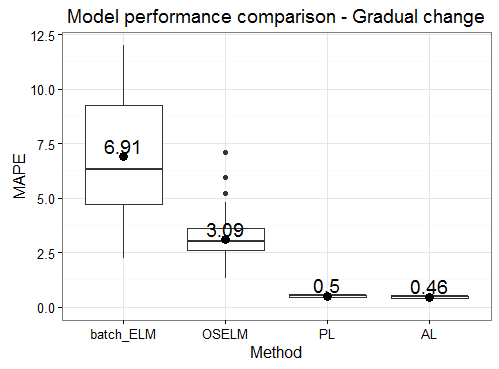}
\caption{The Box-and-Whisker plots of the prediction MAPE when the four algorithms perform on a set of $235$ simulated data streams with {\em gradual}  concept drift. } 
\label{fig-comparison-c}
\end{center}•
\end{figure}

We run the AL algorithm on all of the $500$ simulated dataset and compare the predicted power with the  power output from the GTP simulator and calculate the MAPE measure for each series. Then the same data are modeled by the  three other methods (batch-ELM, OSELM, and PL \footnote[1]{Only the version with the Q registration criterion that leads to a better performance is reported in comparison.}). The key statistics for the resulting distribution  of the  prediction MAPE metric over the $500$ set is summarized in Table \ref{tab}. Figure \ref{fig-comparison-a} and Figure \ref{fig-comparison-c} visually depict the performance comparison of different algorithms under abrupt changes and gradual changes. It is observed that, the framework with dual learners to detect change achieves significantly better prediction accuracy: the average MAPE  for PL and AL are both below $1\%$ and meet industrial standard with standard error of about $0.11\%$, while the average MAPE for the batch-ELM is about $6.799 \pm 3.254 \%$ and for the OSELM is about $3.189 \pm 1.956\%$.  

The poor performance of the batch-ELM can be understood as the concept change exceeds the model generalization capability, in other words, the model  predicts with the initial concept for the entire data stream. The OSELM  adjusts the output weights using \eqref{oselm1}, but the contribution of each new concept data is small comparing with the historical learned information, so it needs more iterations to settle the model. The methods with two learners apply a reset rule so that more aggressively adapt to the up-to-date concept.

\begin{figure}
\begin{center}
\includegraphics[width=.9\columnwidth]{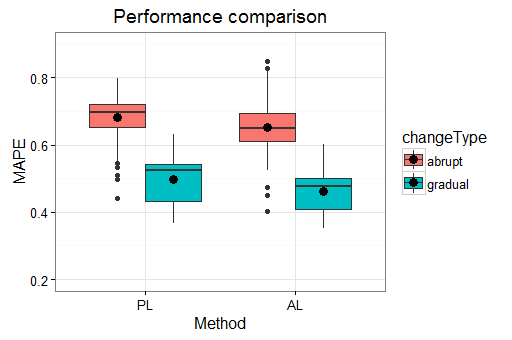}
\caption{The Box-and-Whisker plots of the prediction MAPE when the PL and AL methods are applied on the set of $500$ simulated data streams. } 
\label{fig-plal}
\end{center}•
\end{figure}

The modification of AL upon PL makes it works better on regression setting,  we provide a set of rescaled Box plots in Figure \ref{fig-plal} to better illustrate the improvement.

\begin{table}[!t]
\caption{Summary statistics of MAPE distributions of four algorithms}
\label{tab}
\begin{center}
\begin{tabular}{|c|c|c|c|c|}
\hline
 & batch-ELM & OSELM & PL & AL\\ \hline 
mean & 6.799 & 3.189 & 0.596 & {0.562}\\ \hline
sd & 3.254& 1.956 & 0.111 & 0.112 \\ \hline
\end{tabular}
\end{center}
\end{table}

•

•

•

%\section{Discussion}
%\label{sec:discussion}
%
%\begin{enumerate}
%\item {\bf Modification on the Paired Learner approach:} 
%
% Paired Learner: modification\\
%why least wait,  overfit check, and modifying the alternating condition\item Abrupt vs gradual vs parameter selection
%
%
%\cite{Bach08_pl}
%Figure \ref{fig-plal}
%\end{enumerate}•
%
%The AL framework is developed based on the PL algorithm emphasizing on 

\section{Conclusion}
\label{sec:conclusion}
We proposed an efficient online machine learning framework to model nonstationary data streams with concept drift. By training and maintaining two learners with long and short memory, and compare their prediction performance on new coming batches, the two-model ensemble effectively identifies data or relationship changes. By alternating ensemble prediction output between the two learners, the model continuously adapts to new concept. We performed a numerical algorithm evaluation using simulated power plant operation data. After extensive tests on various change pattens, the proposed algorithm outperforms other baseline and benchmark methods and consistently meets the $1\%$ industrial standard for power performance prediction.

\bibliographystyle{IEEEtran}
\bibliography{ref_ijcnn}
%\bibliography{C:/Users/212428664/Documents/Literature/incremental_learning/bibtex/ref_ijcnn}

\end{document}